%% file: main.tex
\documentclass{article} 

\pdfoutput=1
\usepackage{aloe2022,times}

\input{math_commands.tex}

\usepackage{hyperref}
\usepackage{url}
\usepackage{algorithm}
\usepackage{algorithmicx}
\usepackage{algpseudocode}
\usepackage{graphicx}
\usepackage{wrapfig}

\usepackage{pgfplots} 
\usepgfplotslibrary{fillbetween}
\pgfplotsset{compat=1.17}

\usepackage{subcaption}
\usepackage{booktabs}
\usepackage{enumitem}
\newcommand\std[1]{{\scriptsize \textcolor{darkgray}{$\pm$ #1}}}

\newcommand{\ie}{\textit{i}.\textit{e}., }

\title{
    Walk the Random Walk: Learning to Discover and Reach Goals Without Supervision
}

\author{Lina Mezghani\\
Meta AI \\
Inria \thanks{Univ.\ Grenoble Alpes, Inria, CNRS, Grenoble INP, LJK, 38000 Grenoble, France} \\
\texttt{linamezghani@fb.com} \\
\And
Sainbayar Sukhbaatar \\
Meta AI \\
\And
Piotr Bojanowski \\
Meta AI \\
\And
Karteek Alahari \thanks{Karteek Alahari is supported in part by the ANR grant AVENUE (ANR-18-CE23-0011).} \\
Inria $^*$ \\
}

\iclrfinalcopy 
\begin{document}

\maketitle

\begin{abstract}
Learning a diverse set of skills by interacting with an environment without any external supervision is an important challenge.
In particular, obtaining a goal-conditioned agent that can reach any given state is useful in many applications.
We propose a novel method for training such a goal-conditioned agent without any external rewards or any domain knowledge.
We use random walk to train a \emph{reachability network} that predicts the similarity between two states.
This reachability network is then used in building \emph{goal memory} containing past observations that are diverse and well-balanced.
Finally, we train a goal-conditioned policy network with goals sampled from the goal memory and reward it by the reachability network and the goal memory.
All the components are kept updated throughout training as the agent discovers and learns new goals.
We apply our method to a continuous control navigation and robotic manipulation tasks.


\end{abstract}

\input{intro}
\input{related}
\input{problem}

\input{method}

\input{experiments}

\section{Conclusion}

In this paper, we proposed a novel method for training a goal-conditioned agent without any external supervision.
The method utilizes random walk to learn the similarity between states, which then is used to build a goal memory that is diverse and well-balanced.
On the maze task, we showed that our method can discover increasingly difficult goals.
In this task the graph distance built on the memory worked better than the similarity metric, mainly because the global structure of the rooms is more important than the local dynamics.
The Pusher task showed that our method can learn to manipulate an object without any external supervision.


\bibliography{main}
\bibliographystyle{iclr2022_conference}

\end{document}

%% file: math_commands.tex
\usepackage{amsmath,amsfonts,bm}

\def\eqref#1{equation~\ref{#1}}

\def\1{\bm{1}}

\DeclareMathAlphabet{\mathsfit}{\encodingdefault}{\sfdefault}{m}{sl}
\SetMathAlphabet{\mathsfit}{bold}{\encodingdefault}{\sfdefault}{bx}{n}



%% file: intro.tex
\section{Introduction}
Standard applications of Reinforcement Learning (RL) methods rely on optimizing an objective based on a specific hand-crafted reward function.
However, designing these functions for every specific behaviour that we want the agent to learn is extremely time-consuming, and sometimes not even feasible.
Moreover, optimizing a single reward function restricts the agent to learning a specific behaviour, while it would be more powerful to learn agents that are able to execute diverse skills.
Research efforts have thus been focusing on designing \textit{unsupervised} agents, that are able to learn without external rewards at all.
These agents are usually developed in a two-stage protocol: in the first stage, the agent can interact with the environment and acquire experience without external reward, and in the second stage, it is evaluated on human-designed tasks, with or without adaptation.

The resulting agents are most often trained to explore the environment, and then fine-tuned with few interactions to perform specific tasks.
This adaptation stage still requires a hand-crafted reward function, and the fine-tuned agent is still able to perform only one particular task.
Recently, some works made the link between unsupervised RL and the \textit{goal-conditioned} RL paradigm: in the first stage, the agent would sample goals from its past experience and learn by attempting to reach them, and in the second stage, it would be given user-specified goals to be evaluated on.
This allows for a training protocol where the agent does not need any adaptation to the evaluation task, as the goals are specified in the same way at train and test time, and can execute several behaviours, simply by giving it diverse goals.
However, this paradigm induces several challenges.

First, the agent must have a set of intrinsic goals on which to train.
These goals must be diverse enough, so that the agent learns several behaviours, but not too hard, to make learning feasible.
Previous works rely on past states and sample goals from the replay buffer~\citep{nair2018visual, pong2020skew}, use generative models to create intrinsic goals~\citep{warde2018unsupervised}, or exploit an explorer policy to generate novel goals~\citep{mendonca2021discovering}.
Second, in the absence of supervision, the agent must have an intrinsic reward function shaped for goal-reaching.
Even though this function can be easily hand-crafted in some simple cases like mazes, it can be infeasible in complex real-world environments.
For instance, with high-dimensional inputs like images or complex control tasks such as moving humanoids, assessing whether a state has been achieved can be infeasible.

In a recent line of study, some methods~\citep{hartikainen2019dynamical,venkattaramanujam2019self} proposed to tackle these two challenges by learning a distance network in the state space.
This network can be directly used to shape the intrinsic reward function towards reaching a goal, and can also be used for sampling training goals in a clever way.
These approaches offer a simple and interpretable way of tackling the unsupervised goal-conditioned problem, as the quality of the learned policy depends mostly on the quality of the distance function.

A possible flaw of these methods is that the learning of the distance function and the goal-conditioned policy are intrinsically tied.
Indeed, the distance is learned on samples generated by the policy, and the policy is trained using distance-based rewards.
In contrast to these works, we propose to learn the distance function independently from the goal-conditioned policy, by learning from randomly generated trajectories in the environment.
This self-supervised process, inspired by~\citet{savinov2018episodic}, results in more stability and interpretability when training the distance function jointly with the policy.

%% file: related.tex
\begin{figure}[t]
  \centering
  \includegraphics[width=0.7\linewidth]{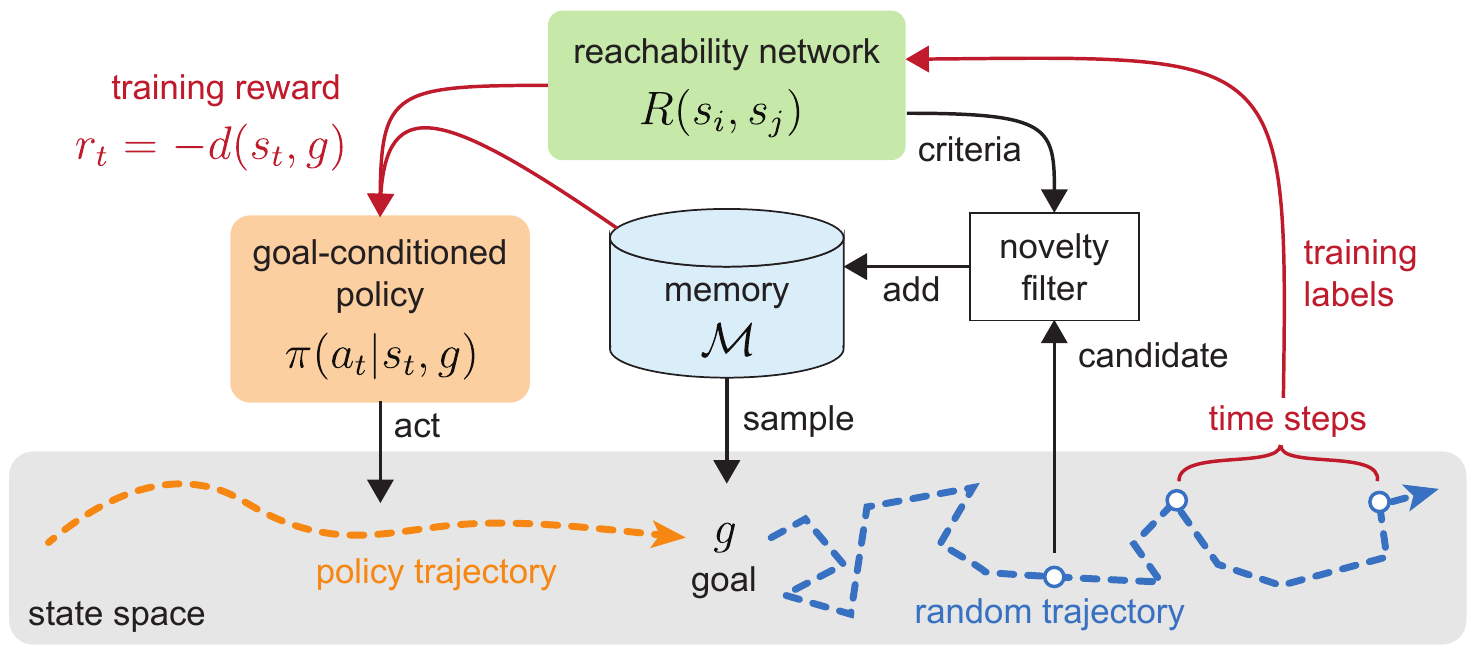}
  \caption{\textbf{Overview of our method.} 
  The agent performs two consecutive stages.
  In the first stage (orange arrow), it samples a goal $g$ from the memory $\mathcal{M}$ and executes a trajectory with policy $\pi$ towards a goal with reward $r_t = - d(s_t, g)$.
  In the second stage (blue arrow), the agent performs random steps in order to discover novel goals to be added to the memory $\mathcal{M}$, and to generate training data for the Reachability Network $R$.}
  \label{fig:method}
\end{figure}

\section{Related Work}

In its original formulation, goal-conditioned reinforcement learning was tackled by several methods~\citep{kaelbling1993learning, schaul2015universal, andrychowicz2017hindsight, nasiriany2019planning}.
The policy learning process is supervised in these works: the set of evaluation goals is available at train time as well as a shaped reward function that guides the agent to the goal.

Several works propose solutions for generating goals automatically when training goal-conditioned policies, including self-play~\citep{sukhbaatar2018intrinsic, sukhbaatar2018learning, openai2020asymmetric}, where an agent learns to reach goals with an adversarial objective and a second agent that proposes goals of increasing difficulty.
In the same spirit, \cite{campero2020learning} presents a student-teacher policy trained in a single module with an adversarial learning loss.
These methods assume access to a hand-crafted goal achievement function and therefore require supervision.

In a recent line of research, some works~\citep{nair2018visual, pong2020skew, warde2018unsupervised, mendonca2021discovering} focused on learning goal-conditioned policies in an unsupervised fashion.
In these works, the objective is to train general agents that can reach any goal state in the environment without any supervision (reward, goal-reaching function) at train time.
There are two challenges that these methods have to overcome: first, how to generate goals at train time, and second how to assess whether a goal was reached or not.

To tackle these challenges, \cite{warde2018unsupervised} propose to learn a goal achievement reward function jointly with the goal-conditioned policy with a mutual information objective.
\cite{nair2018visual} train a variational auto-encoder and generate goals in its latent space while using the euclidean distance in this space to compute a dense reward function.
\cite{pong2020skew} improves this approach by ``skewing" the set of goals to encourage exploration.
More recently, \cite{mendonca2021discovering} presented a model-based method, composed of an explorer, which proposes novel goals in the latent space and an achiever that learns to reach these goals.

Closer to our work, \cite{venkattaramanujam2019self} and \cite{hartikainen2019dynamical} learn a distance function in the state space jointly with the goal-conditioned policy.
This distance function is used to compute rewards for the policy, but also to sample goals in clever ways: by encouraging far-away goals~\citep{hartikainen2019dynamical} or by sampling goals of intermediate difficulty~\citep{venkattaramanujam2019self}.
Our work extends this line of research with a new way of learning the distance function: independently from the policy by training it on trajectories sampled by a random policy at every stage of the policy learning process.

%% file: problem.tex
\section{Problem Formulation}
\label{sec:problem}

In the classical RL setup, 
the agent observes a state $s_t \in \mathcal{S}$ at time $t$, selects action 
$a_t$ 
according to a policy $\pi(a_t|s_t)$ and receives a reward $r_t \in \mathbb{R}$.
The agent learns by maximizing the cumulative return $ \sum_{t=1}^{T} r_t$ where $T$ is the episode length.
This classical RL setup relies on the design of a hand-crafted reward function to solve a specific task.

In this work, however, we are interested in learning agents that have diverse behaviours, and are not limited to mastering a single task.
Given a set of target states $G_{eval} \subseteq \mathcal{S}$, the agent is evaluated on its ability to achieve these goals.
We assume $G_{eval}$ is not known during training, so our objective is to learn a goal-conditioned policy $\pi(a_t|s_t, g)$ capable of reaching any goal state $g \in \mathcal{S}$.



%% file: method.tex
\section{Method}

\input{algo}

We propose a novel method to tackle the problem of unsupervised training of a goal-conditioned agent.
Our method comprises three components as shown in \autoref{fig:method}.
The first component is a reachability network (RNet) that learns to predict the similarity between any given two states.
We train this RNet using random walk trajectories generated by a random policy. 
The second component is a goal memory $\mathcal{M}$ that stores previously seen states that are diverse. 
To ensure this diversity, we employ the RNet as a criterion to avoid adding similar states to $\mathcal{M}$. 
The last component is a goal-conditioned policy $\pi(a_t|s_t, g)$ that is trained to reach goals $g$ sampled from $\mathcal{M}$.
The policy is trained with rewards $r_t=-d(s_t, g)$ using a distance metric.
Since we do not have access to a hand-crafted distance metric, we propose two ways to approximate it using the RNet and $\mathcal{M}$. 

The important aspect of our method is continued exploration of an environment that allows the agent to discover new areas.
This exploration is accompanied by gradual progression from easy goals to increasingly difficult goals, which acts as curriculum for better training of the policy.
This is achieved by first starting an episode with the policy $\pi$ acting towards a goal sampled $g \in \mathcal{M}$ as shown by the orange trajectory in Figure~\ref{fig:method}.
Once the policy trajectory ends, we start a random walk (blue trajectory), which is likely to cover areas that are beyond the reach of the current policy.
As both RNet and the goal memory $\mathcal{M}$ are updated by those random trajectories, the agent will be discovering new goals that are sufficiently different from existing ones.
This, in turn, will train the policy to reach increasingly further away goals.
Since this discovery of increasingly harder goals need to continue throughout the training, all three components are continuously updated as shown in \autoref{alg:method}. 
Let us now describe each component in more detail and how they interact with each other.

\subsection{Reachability Network}
To learn the similarity between states from random trajectories, we train a model similar to the Reachability Network, which was first introduced by \citet{savinov2018episodic}.
The idea introduced in this work is to approximate the distance between states in the environment by the average number of steps it takes for a random policy to go from one state to another.

During training, we collect random trajectories $(s^a_1, ..., s^a_N)$ in \emph{random buffer} $\mathcal{B}$ where $a$ is a trajectory index.
We define a reachability label $y^{ab}_{ij}$ for each pair of observations $(s^a_i, s^b_j)$ that depends on their distance in the sequence and if $a$, $b$ are the same trajectory.
More precisely,
\begin{equation}
	y^{ab}_{ij} = \begin{cases}
			1 \quad \text{if} \ a = b \ \text{and} \ |i - j| \leq \tau_\text{reach}, \\
			0 \quad \text{otherwise},
		 \end{cases}
	\quad \text{for } 1 \leq i, j \leq T 
\end{equation}
where the \textit{reachability threshold} $\tau_\text{reach}$ is a hyperparameter.
We train a siamese neural network $R$, the RNet, to predict the reachability label $y^{ab}_{ij}$ from a pair of observations $(s^a_i, s^b_j)$ in $\mathcal{B}$. 
The RNet consists of an embedding network $g$, and a fully-connected network $f$ to compare the embeddings, \ie
\begin{equation}
	R(s^a_i, s^b_j) = \sigma \left[ f(g(s^a_i), g(s^b_j)) \right],
\end{equation}
where $\sigma$ is a sigmoid function.
A higher $R$ value will indicate that two states easily reachable with random walk, so can be considered close in the environment.
The training of the RNet is self-supervised, as the supervised labels needed to train the network are automatically generated.

\subsection{Goal Memory}
\label{sec:memory}

Since the set of evaluation goals $G_{eval}$ is not known at train time, the agent needs to come up with goals on which to train on for itself.
One possible solution would be to sample goals from states that the agent has previously seen (\ie from the replay buffer).
The main downfall of this solution is that there is no incentive in discovering novel states, and the agent would potentially learn to reach only a small proportion of the state space.
Instead, we propose to incrementally build a set of intrinsic goals, or \emph{goal memory} $\mathcal{M}$ on which the agent trains on.
We used the random trajectories $\mathcal{B}$ to build $\mathcal{M}$ rather than the policy trajectories because the random walks are more likely to contain novel states.

\paragraph{Memory filtering:} The states in $\mathcal{B}$ have to go through a filter: a state is added to the memory $\mathcal{M}$ only if it is distant enough from all other goals in $\mathcal{M}$.
More precisely, a state $s \in \mathcal{B}$ is added to $\mathcal{M}$ if and only if $\forall m \in \mathcal{M}, R(s, m) < \tau_\text{memory}$, where the \emph{memory threshold} $\tau_\text{memory}$ is a hyperparameter.
This filtering avoids redundancy by preventing similar states to be added to the memory.
It also has a balancing effect because it limits the number of goals that can be added from a certain area even if it is visited by the agent many times.
This is especially important if episodes always start from the same initial state $s_0$ and most samples in $\mathcal{B}$ are concentrated around $s_0$. 

\paragraph{Weighted goal sampling:}
Optionally, the same balancing effect as the filtering can be achieved by giving appropriate sampling weights to the goals.
Let us consider a state $s_i$ in the memory $\mathcal{M}$
and its \emph{reachable area} defined as $A_i=\{s_j | R(s_i, s_j) >0 \}$.
The filtering ensures that there is only one state in $\mathcal{M}$ from $A_i$, so the probability of sampling a goal from $A_i$ is the same as any other reachable areas.
While this keeps the memory more balanced and evenly distributed, it also limits the number of goals we train on, and can make the policy overfit to these goals.
So instead, we can use a weighting scheme that will allow $k$ states from $A_i$ to be added to $\mathcal{M}$.
In turn, goals from $A_i$ are sampled with a probability proportional to $1/k$ so each reachable area is still equally represented in the sampled goals.

\subsection{Distance function for policy training}

When an appropriate distance function $d$ in the state space is available, designing a reward function to learn a goal-conditioned policy towards a goal $g$ is straightforward, by setting $r_t = - d(s_t, g)$.
However, in realistic environments with complex high-dimensional inputs like images, designing a distance between states is generally not feasible.
We therefore assume that we do not have access to any kind of distance in the state space \emph{a priori}.
Instead, we propose two different ways to construct a distance function.

\paragraph{RNet distance:}
The RNet predicts the similarity between $s_i$ and $s_j$ so we can directly use it as a distance metric.
However, the RNet has a sigmoid function $\sigma$ for binary classification. We remove it, and define a distance metric as $d(s_i, s_j) = - f(e(s_i), e(s_j))$.

\paragraph{Graph distance:}
Since the RNet is trained to determine whether states are close to each other or not, there is no guarantee that the aforementioned RNet distance will have good global properties.
Instead of using directly the RNet, we construct an unweighted graph on the memory, whose nodes are the memory states, and edges are between nodes that have high reachability score. More precisely, the memory graph contains an edge between states $s_i$ and $s_j$ if and only if $R(s_i, s_j) > \tau_\text{graph}$, where $\tau_\text{graph}$, the \textit{graph threshold}, is a hyperparameter.

Using this memory graph, we can easily derive a distance function $d$ between any pair of states in $\mathcal{M}$ by computing the length of the shortest path in this graph, providing that the graph is connected.
Moreover, we can extend this distance to all states in the state space $\mathcal{S}$ by computing, for a state $s \in \mathcal{S}$, its closest node $n_s$ in the memory \textit{w.r.t} the RNet \textit{i.e.} $n_s = \underset{m \in \mathcal{M}}{\mathrm{argmax}} \quad R(s, m)$.
The distance $d$ between two states in the state space becomes the length of the shortest path between their respective closest nodes in the graph.
This process allows for propagating the good local properties of the RNet in order to get a well-shaped distance function for faraway states.

%% file: algo.tex
\begin{algorithm}[tb]
\footnotesize
  \caption{Training of our method}
  \label{alg:method}
\begin{algorithmic}
    \State \textbf{Initialize:} random buffer $\mathcal{B} \gets \emptyset$, goal memory $\mathcal{M} \gets \emptyset$
    \State \textbf{Warm-up:} collect random trajectories and store them in $\mathcal{B}$
    \For{each training stage}
        \State Train the reachability network $R$ on trajectories from $\mathcal{B}$
        \State Update the goal memory $\mathcal{M}$ with states from $\mathcal{B}$ using $R$ as a criteria. 
        \For{each episode}
            \State Sample a goal $g$ from the memory $\mathcal{M}$
            \State Run the policy $\pi(a_t | s_t, g)$ for $T$ steps and train it with rewards $r_t=-d(s_t, g)$
            \State Take $N$ random steps and add that trajectory to $\mathcal{B}$
        \EndFor
    \EndFor

\end{algorithmic}
\end{algorithm}

%% file: experiments.tex
\begin{figure}[t]
\begin{subfigure}{.24\linewidth}
    \centering
    \includegraphics[width=.8\linewidth]{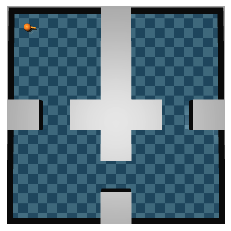}
    \caption{Maze environment}
    \label{fig:maze_perf_a}
\end{subfigure}
\begin{subfigure}{.25\linewidth}
    \centering
    \includegraphics[width=\linewidth]{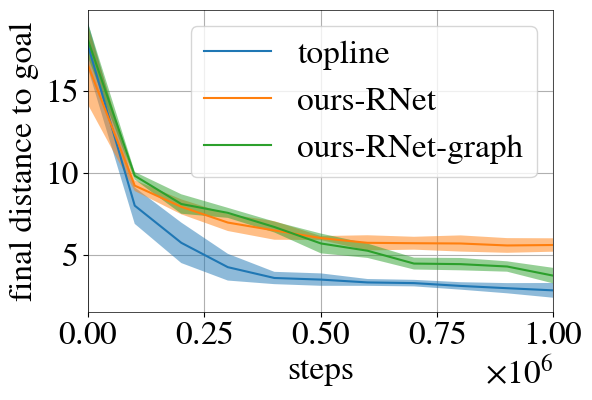}
    \caption{Evaluation performance}
    \label{fig:maze_perf_b}
\end{subfigure}
\begin{subfigure}{.24\linewidth}
    \centering
    \includegraphics[width=\linewidth]{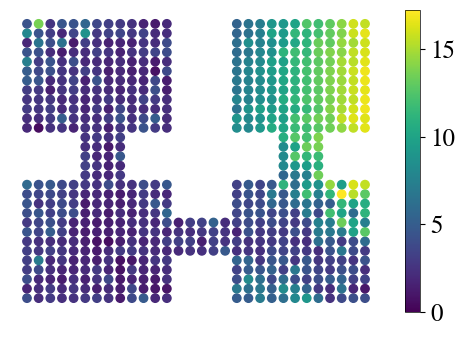}
    \caption{final distance to goal - \textbf{ours-RNet}}
    \label{fig:maze_perf_c}
\end{subfigure}
\begin{subfigure}{.24\linewidth}
    \centering
    \includegraphics[width=\linewidth]{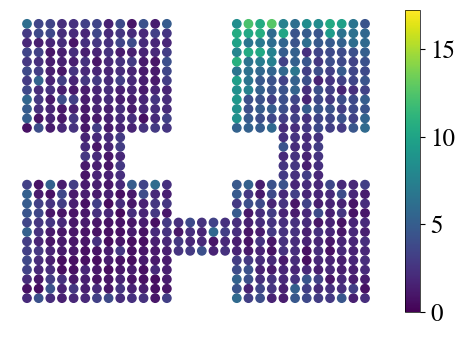}
    \caption{final distance to goal - \textbf{ours-RNet-graph}}
    \label{fig:maze_perf_d}
\end{subfigure}
    \caption{Performance on the Maze environment.}
    \label{fig:maze_perf}
\end{figure}

\section{Experiments}
We test our method on two continuous control tasks: maze navigation and robotic arm manipulation. For the policy training, we use Soft actor-critic \citep{haarnoja2018soft}. The policy training steps are limited to 1M steps.
For each model, we report mean and standard deviation over 5 random seeds.

\subsection{Maze environment}

We first evaluate our model on a simple maze environment~\citep{mujocomaze} with four rooms, as shown in \autoref{fig:maze_perf_a}.
The point agent starts always at the same position, in the top left corner, and can move in the maze by performing actions in a continuous space.
Here, the observations are state-based: they contain the agent's position, direction and velocity.
We generate an evaluation set of 500 goals sampled at random in the environment, and we assess the performance of the agent by measuring the distance between its final position and the goal position.

We compare the following three models:
\textbf{(i) topline} the fully-supervised topline: the agent samples goal uniformly at random in the evaluation set at train time and uses the euclidean distance in the maze for dense reward computation, \textbf{(ii) ours-RNet} our method with rewards computed directly from the RNet distance, and \textbf{(iii) ours-RNet-graph} our method with rewards computed as the shortest path in the memory graph.

\paragraph{Evaluating the performance of the model}
We first show the evaluation performance of the 3 models during training in \autoref{fig:maze_perf_b}.
We see that the unsupervised model \textbf{ours-RNet} performs significantly worse than the supervised one (\textbf{topline}), but that the gap is largely reduced by using graph-based rewards (\textbf{ours-RNet-graph}).

We then perform a qualitative evaluation of the final distance to the goal for the two different rewards in \autoref{fig:maze_perf_c} and~\ref{fig:maze_perf_d}.
It shows that the policy trained with graph rewards is able to reach almost all goals in the environment, including the ones that are in the fourth room, while the model with the RNet rewards is not able to achieve them.

\begin{figure}[t]
\begin{subfigure}{.33\linewidth}
    \centering
    \includegraphics[width=\linewidth]{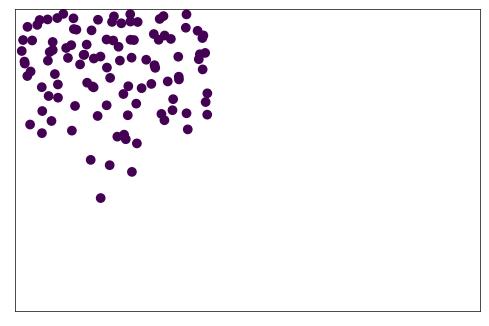}
    \caption{step = 0}
\end{subfigure}
\begin{subfigure}{.33\linewidth}
    \centering
    \includegraphics[width=\linewidth]{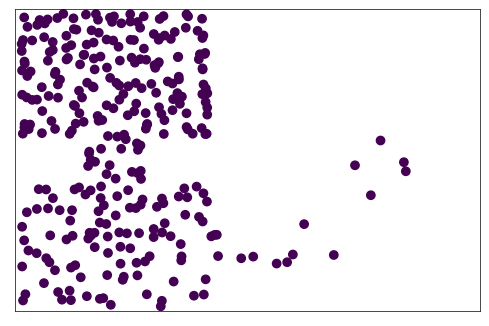}
    \caption{step = 200k}
\end{subfigure}
\begin{subfigure}{.33\linewidth}
    \centering
    \includegraphics[width=\linewidth]{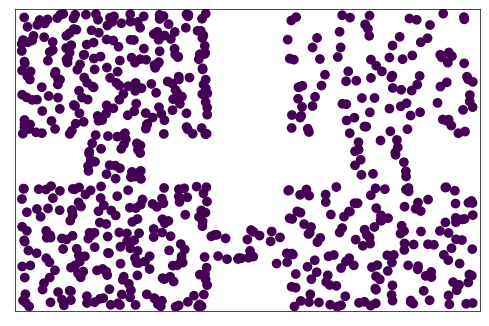}
    \caption{step = 1M}
\end{subfigure}
\caption{
        Memory of the \textbf{ours-RNet-graph} model during training.
}
    \label{fig:maze_mem_training}
\end{figure}

\paragraph{Analysis of the Reachability Network}
In order to understand how our unsupervised agent learns to discover and achieve novel goals, we visualize the memory $\mathcal{M}$ at several steps of training in \autoref{fig:maze_mem_training}. We see that throughout training, the memory gets filled with vectors that are further and further away from the initial state, and that after 1M steps, it is well distributed in the state space.

\begin{figure}[t]
\centering
\begin{subfigure}{.35\linewidth}
    \centering
    \includegraphics[width=\linewidth]{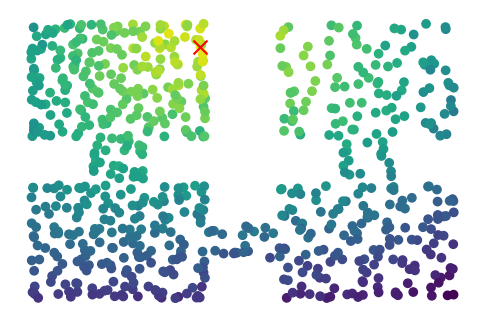}
    \caption{RNet distance}
\end{subfigure}
\begin{subfigure}{.35\linewidth}
    \centering
    \includegraphics[width=\linewidth]{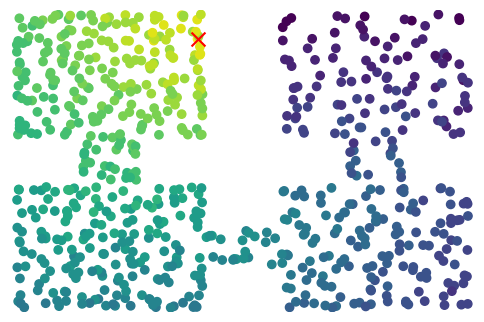}
    \caption{Graph distance}
\end{subfigure}
    \caption{
        Visualization of rewards computed with the RNet (a) and graph (b) distances.}
    \label{fig:graph_dist}
\end{figure}

We then compare the shape of the reward computed from the RNet reward function, and using the graph memory in \autoref{fig:graph_dist}.
We see that the graph-based reward is necessary in order to have a reward function that takes into account the true dynamics of the environment.
Indeed, the shape of the reward based on the RNet distance will make the agent bump into the wall to go from room 1 to room 4, while the graph-based reward has a smooth shape that follows the room order.

\begin{figure}[t]
\begin{subfigure}{.24\linewidth}
    \centering
    \includegraphics[width=\linewidth]{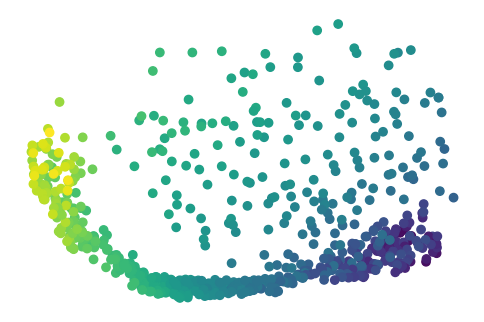}
    \caption{step = 0}
    \label{fig:maze_emb_a}
\end{subfigure}
\begin{subfigure}{.24\linewidth}
    \centering
    \includegraphics[width=\linewidth]{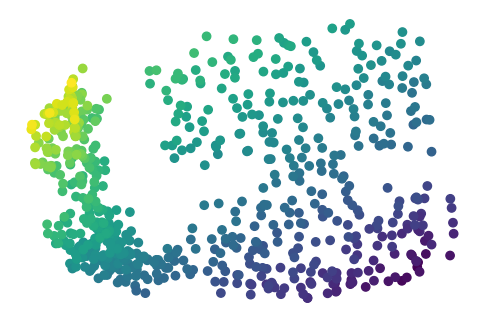}
    \caption{step = 200k}
    \label{fig:maze_emb_b}
\end{subfigure}
\begin{subfigure}{.24\linewidth}
    \centering
    \includegraphics[width=\linewidth]{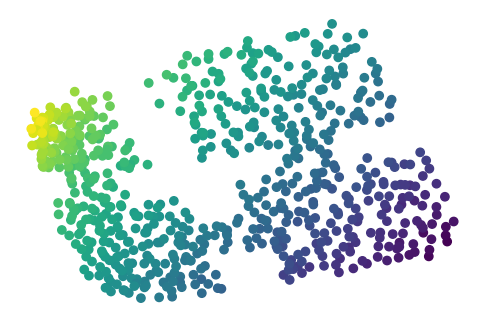}
    \caption{step = 500k}
    \label{fig:maze_emb_c}
\end{subfigure}
\begin{subfigure}{.24\linewidth}
    \centering
    \includegraphics[width=\linewidth]{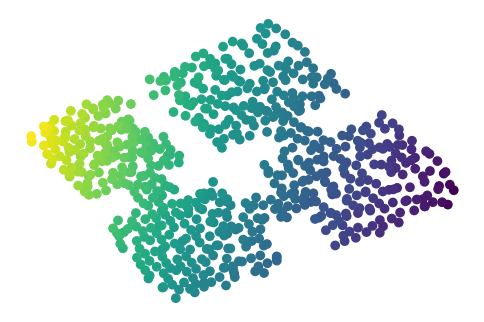}
    \caption{step = 1M}
    \label{fig:maze_emb_d}
\end{subfigure}
    \caption{
        Visualization of the RNet embedding $g(s)$ for the \textbf{ours-RNet-graph} model. 
        We compute the 2D-PCA for a set of points sampled in the environment at several steps of training.
    }
    \label{fig:maze_emb}
\end{figure}

Finally, we visualize the features learned by the RNet embedding for the \textbf{ours-RNet-graph} model in \autoref{fig:maze_emb}.
We generate a set of points in the environment, and compute their features with the embedding part of the RNet.
We then reduce the dimension of the embeddings by computing the 2D-PCA and visualize the resulting projection.
We see that throughout training, the RNet learns to disentangle observations in consecutive rooms.
After 1M steps, the embeddings describe well the entire state space.

\subsection{Pusher Task}

\begin{figure}[t]
    \centering
    \includegraphics[width=0.3\linewidth]{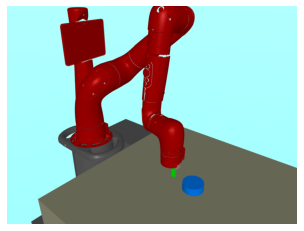}
    \hskip 10pt
    \input{fig/pusher_converge}
    \caption{\textbf{Left:} Pusher environment. The robot arm has to move the puck to a specified location. \textbf{Right:} Learning curve on the Pusher-Vec environment showing the distance to the goal.}
    \label{fig:pusher_env}
\end{figure}

Next, we apply our method to a realistic robotic environment from \cite{nair2018visual}. In particular, we use the \emph{Pusher} task shown in \autoref{fig:pusher_env}~(left) where a robot arm (red) needs to push a puck (blue) to a specified location on a table. 
The performance of this task is measured by the final euclidean distance $d^*(s_t, g)$ between the puck and its target location.
See \cite{nair2018visual} for more details about the environment.

We experiment with two versions of this environment: \emph{Pusher-RGB} where observations are RGB images, and \emph{Pusher-Vec} where observations are a vector containing the hand and puck locations.
Even with the vector observations, this task is challenging because our method is not given any information particular to this task such as the importance of moving the puck, or the distance between the puck and its target location. Instead, our method has to learn all this solely from interacting with the environment without any external reward.

We start with \emph{Pusher-Vec} and compare our model against a supervised \emph{topline} that is rewarded by the oracle distance $r_t=-d^*(s_t, g)$.
As shown in \autoref{fig:pusher_env}~(right), the topline learns quickly to move the puck closer to the target location.
Our method also learns to move the puck to the target location, 
with the RNet reward working slightly better than the graph reward. For all remaining experiments, we use the RNet reward.

Our method does not know that moving the puck is the goal. Instead, it tries to match the current state to the target state according to its own distance metric.
In fact, we can see the effect of this if we look at the hand distance.
The target observation contain a hand location, but it has no effect on the evaluation metric.
While the topline has no incentive to match the hand location, our method also learned to match the hand position (hand distance 0.08 vs 0.02), which can be a useful skill in general.

We compare our method against two existing baselines: Skew-Fit~\citep{pong2020skew} and LEXA~\citep{mendonca2021discovering} on the Pusher-RGB task in \autoref{tab:pusher_final}.
We evaluated our model on the same set of evaluation goals than these methods, and reported the performance available in their paper.
We see that our method performs better the existing model-free method Skew-Fit, but is still far behind the model-based model LEXA.
One possible explanation for this bad performance is the quality of the RNet on RGB, which could be greatly improved to match Pusher-Vec performance.

\begin{figure}[t]
\begin{subfigure}{.24\linewidth}
    \centering
    \includegraphics[width=\linewidth]{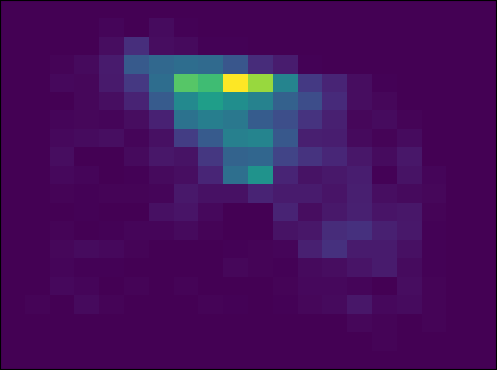}
    \caption{step=160k}
    \label{fig:pusher_buffer_1}
\end{subfigure}
\begin{subfigure}{.24\linewidth}
    \centering
    \includegraphics[width=\linewidth]{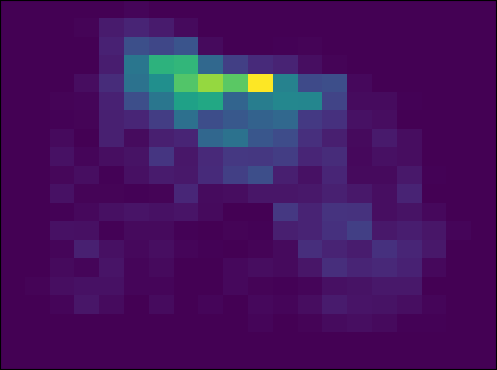}
    \caption{step=331k}
\end{subfigure}
\begin{subfigure}{.24\linewidth}
    \centering
    \includegraphics[width=\linewidth]{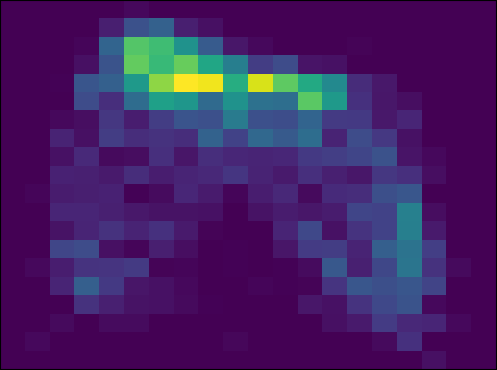}
    \caption{step=673k}
\end{subfigure}
\begin{subfigure}{.24\linewidth}
    \centering
    \includegraphics[width=\linewidth]{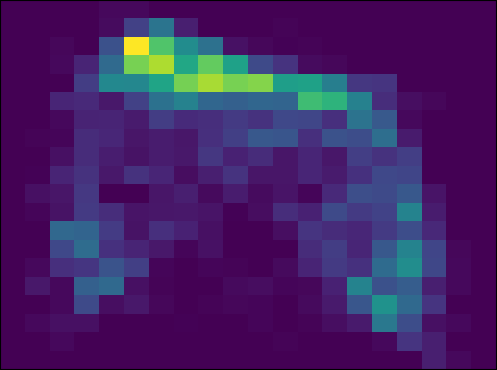}
    \caption{step=995k}
\end{subfigure}
    \caption{
        Distribution of the puck locations in random buffer $\mathcal{B}$ at different stages of training.
    }
    \label{fig:pusher_buffer}
\end{figure}

\subsubsection{Effectiveness of the reachability network}

\begin{figure}[t]
\begin{subfigure}{.39\linewidth}
    \centering
    \input{fig/pusher_result}
    \caption{}
    \label{fig:pusher_oracle}
\end{subfigure}
\begin{subfigure}{.3\linewidth}
    \centering
    \includegraphics[width=0.8\linewidth]{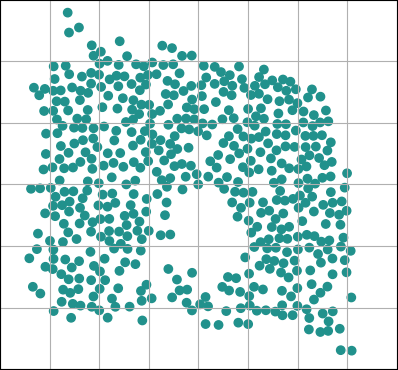}
    \caption{Oracle distance}
\end{subfigure}
\begin{subfigure}{.3\linewidth}
    \centering
    \includegraphics[width=0.8\linewidth]{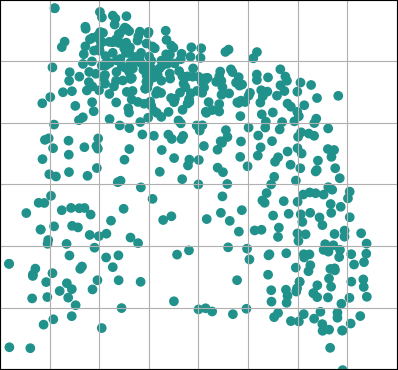}
    \caption{Reachability network}
\end{subfigure}
\caption{Ablation on Pusher-Vec where the reachability network is replaced by the oracle puck  distance. (a) Performance comparison against the three oracle variations. (b,c) Puck positions of the states stored in the memory.}
\label{fig:pusher_memory}
\end{figure}

The RNet $R(s_i,s_j)$ is an integral part of our method and it has to function well for our method to succeed.
Here, we test the quality of the RNet with an ablation study. 
We replaced the RNet in our method with the oracle distance metric $d^*(s_i, s_j)$.
Since our method rely on $R(s_i,s_j)$ in two places, when giving a reward to the policy and as a criteria for adding to the memory, we ended up with three oracle variations: 
\begin{itemize}[noitemsep,topsep=0pt,parsep=0pt,partopsep=0pt]
    \item \textbf{oracle-reward:} the oracle distance is used as a reward $r_t=-d^*(s_t, g)$ for the policy.
    \item \textbf{oracle-memory:} the oracle distance is used as a criteria for adding a state to the memory. A state $s$ is added when $d^*(s, m) > 0.0075$ for $\forall m \in \mathcal{M}$.
    \item \textbf{oracle:} the oracle distance is used for both, replacing the RNet completely.
\end{itemize}
The results are shown in \autoref{fig:pusher_oracle} where we do not see much difference between our method and its oracle variations in terms of performance. This indicates that the RNet is as effective as the oracle distance metric.

However, the RNet predictions differ from the oracle distance, which can be seen from the states stored in the memory.
\autoref{fig:pusher_memory}b,c shows the puck positions of the states that are stored in the memory. The oracle distance is based on the exact puck position, so the memory looks more uniform and evenly spaced.
This is not true when the RNet is used, which takes into account the hand position as well. It also depends on the agent's experience, so the memory is more biased towards puck locations that are relatively easy to achieve.

\subsubsection{The Effect of The Memory Filtering}
As discussed in Sec.~\ref{sec:memory}, one goal of filtering states added to the memory is to make the memory more evenly distributed over explored areas.
To measure the effect of the filtering, we trained several variations of our method. First, we made the filter loose by increasing its threshold $\tau_\text{memory}$ from its default value of 0.5, to allow more states to be added to the memory.
This clearly had a negatively impact on the performance as shown in the top 3 rows of \autoref{tab:filter}.
Removing the filter altogether, as shown in the 4th row, leads to the worst performance.
Next, we use the weighting scheme from Sec.~\ref{sec:memory} that take on the role of the filtering and keep the goals well balanced. When it is applied to a loose filtering threshold of 0.95, it improved the performance and actually gave the best performance we obtained on Pusher-Vec.
The reason why the weighting worked better than the filtering alone is probably because it had more goals that made overfitting less likely.

\begin{table}[t]
\parbox{.6\linewidth}{
    \centering
    \footnotesize
    \begin{tabular}{cccc}
        \toprule
        Memory      & Threshold    & Goal          & Goal distance \\
        Filter      & $\tau_\text{memory}$ & weighting     & ($\cdot 10^{-2}$) \\ \midrule
        \checkmark  & 0.5         & $\times$      & 2.05 \std{0.21} \\
        \checkmark  & 0.73         & $\times$      & 2.47 \std{0.84} \\
        \checkmark  & 0.95         & $\times$      & 5.75 \std{2.27} \\
        $\times$    & -         & $\times$      & 6.04 \std{2.32} \\
        \checkmark  & 0.95         & \checkmark    & 1.77 \std{0.21} \\
        \bottomrule
    \end{tabular}
    \caption{Effect of the memory filtering and goal weighting on the Pusher-Vec performance.}
    \label{tab:filter}
}
\hfill
\parbox{.35\linewidth}{
    \centering
    \footnotesize
    \begin{tabular}{lc}
    \toprule
        Method & Goal distance \\
         & ($\cdot 10^{-2}$) \\ \midrule
        Skew-fit & 4.9 \\
        LEXA & 2.3\\ \midrule
        Topline & 1.34 \std{1.09} \\
        Our method & 4.11 \std{0.53} \\
        \bottomrule
    \end{tabular}
    \caption{The performance on Pusher-RGB}
    \label{tab:pusher_final}
}
\end{table}

%% file: fig/pusher_converge.tex
\begin{tikzpicture}[every node/.style={scale=0.8}]


\begin{axis}[
    width=7cm, height=4cm,
    ylabel = {distance to goal},
    xlabel = {policy training steps},
    grid=major,
    enlarge x limits=0,
    legend entries={Topline,Ours-RNet, Ours-RNet-graph},
    legend style={cells={anchor=west}},
]

\addplot+[
    blue,
    mark=none,
] table [x=step,y=topline-mean]{fig/pusher_converge.dat};

\addplot+[
    red,
    mark=none,
] table [x=step,y=unsup-mean]{fig/pusher_converge.dat};

\addplot+[
    orange,
    mark=none,
] table [x=step,y=graph-mean]{fig/pusher_converge.dat};

\addplot[
    name path=upper,
    draw=none,
] table [x=step,y expr=\thisrow{topline-mean}+\thisrow{topline-std}]
    {fig/pusher_converge.dat};
\addplot[
    name path=lower,
    draw=none,
] table [x=step,y expr=\thisrow{topline-mean}-\thisrow{topline-std}]
    {fig/pusher_converge.dat};
\addplot [fill=blue, fill opacity=0.1] fill between[of=upper and lower];

\addplot[
    name path=upper,
    draw=none,
] table [x=step,y expr=\thisrow{unsup-mean}+\thisrow{unsup-std}]
    {fig/pusher_converge.dat};
\addplot[
    name path=lower,
    draw=none,
] table [x=step,y expr=\thisrow{unsup-mean}-\thisrow{unsup-std}]
    {fig/pusher_converge.dat};
\addplot [fill=red, fill opacity=0.1] fill between[of=upper and lower];

\addplot[
    name path=upper,
    draw=none,
] table [x=step,y expr=\thisrow{graph-mean}+\thisrow{graph-std}]
    {fig/pusher_converge.dat};
\addplot[
    name path=lower,
    draw=none,
] table [x=step,y expr=\thisrow{graph-mean}-\thisrow{graph-std}]
    {fig/pusher_converge.dat};
\addplot [fill=orange, fill opacity=0.1] fill between[of=upper and lower];

\end{axis}

\end{tikzpicture}

%% file: fig/pusher_result.tex
\begin{tikzpicture}[every node/.style={scale=0.8}]

\footnotesize

\begin{axis}[
    width=6cm, height=4cm,
    ylabel = distance to goal,
    ybar,
    bar width=15pt,
    xtick=data,
    grid=major,
    symbolic x coords={topline,oracle,oracle reward,oracle memory,unsup},
    xticklabel style={text width=1.3cm, align=center},
    ymin=0, ymax=0.05,
    enlarge x limits=0.15,
]

\addplot+[
    cyan!40!black,
    fill=cyan!20!white,
    error bars/.cd,
        y dir=both,y explicit,
] table [x=x,y=y,y error=err,col sep=comma] {
    x,                  y,          err
    oracle,             0.0225,     0.0164
    oracle reward,      0.0124,     0.0015
    oracle memory,      0.0129,     0.0006
    unsup,              0.0204,     0.0024
};

\end{axis}

\end{tikzpicture}